\documentclass[sigconf]{acmart}
\usepackage{graphicx}

\AtBeginDocument{%
  \providecommand\BibTeX{{%
    \normalfont B\kern-0.5em{\scshape i\kern-0.25em b}\kern-0.8em\TeX}}}

\setcopyright{rightsretained}
\copyrightyear{2021}
\acmYear{2021}
\acmDOI{}
\acmISBN{} 

\begin{document}

\title[Transfer learning to study burned area in Northern Uganda]{Using transfer learning to study burned area dynamics: A case study of refugee settlements in West Nile, Northern Uganda}

\author{Robert Huppertz}
\email{prhuppertz@gmail.com}
\affiliation{%
  \institution{NASA Harvest}
  \country{USA}
}
\author{Catherine Nakalembe}
\orcid{0000-0002-2213-593X}
\affiliation{%
  \institution{University of Maryland}
  \streetaddress{2181 Lefrak Hall}
  \city{College Park}
  \state{MD}
  \country{USA}
  \postcode{20740}}
 \affiliation{%
  \institution{NASA Harvest}
  \country{USA}
}
\author{Hannah R. Kerner}
\orcid{0000-0002-3259-7759}

\affiliation{%
  \institution{University of Maryland}
  \streetaddress{2181 Lefrak Hall}
  \city{College Park}
  \state{MD}
  \country{USA}
  \postcode{20740}}
  
 \affiliation{%
  \institution{NASA Harvest}
  \country{USA}
}
  
\author{Ramani Lachyan}
\affiliation{%
  \institution{Cervest}
  \city{London}
  \country{UK}}
  
\author{Maxime Rischard}
\affiliation{%
  \institution{Cervest}
  \city{London}
  \country{UK}}

\renewcommand{\shortauthors}{Huppertz, et al.}

\begin{abstract}
With the global refugee crisis at a historic high, there is a growing need to assess the impact of refugee settlements on their hosting countries and surrounding environments. Because fires are an important land management practice in smallholder agriculture in sub-Saharan Africa, burned area (BA) mappings can help provide information about the impacts of land management practices on local environments. However, a lack of BA ground-truth data in much of sub-Saharan Africa limits the use of highly scalable deep learning (DL) techniques for such BA mappings. In this work, we propose a scalable transfer learning approach to study BA dynamics in areas with little to no ground-truth data such as the West Nile region in Northern Uganda. We train a deep learning model on BA ground-truth data in Portugal and propose the application of that model on refugee-hosting districts in West Nile between 2015 and 2020. By comparing the district-level BA dynamic with the wider West Nile region, we aim to add understanding of the land management impacts of refugee settlements on their surrounding environments.
\end{abstract}



\maketitle

\section{Introduction}
The global refugee crisis is at a historic high with a staggering 26 million refugees and 79.5 million forcibly displaced people globally, as of end-2019 \cite{UNHCR2020}. Uganda hosts the largest refugee population in Africa and the fourth largest in the world. The country has a long history of hosting refugees dating back to the late 1950s and has experienced continuous flows of refugees from Rwanda, Burundi, Ethiopia, the Democratic Republic of the Congo, and South Sudan that are fleeing from persisting violence in their home countries \cite{Mulumba2014}. Many new refugee settlements have been established in the West Nile region of Northern Uganda, following renewed conflict particularly in South Sudan since 2013 \cite{Ahimbisibwe2019}. Although refugee movements and impacts on hosting countries have been studied for a long time, there is little understanding of how refugee settlements impact the local environment they are forcibly displaced to and the long-term consequences of this \cite{Nakalembe2021,hoek2017}. In this regard, the Global Compact on Refugees (GCR) was established to ease the burden of those refugee settlements on hosting countries while emphasizing the importance of better insights into the refugee-environment nexus \cite{Maystadt2020}. 

While there is already a lack of understanding of the refugee-environment nexus globally, this is further complicated by the changing dynamics of local environmental conditions through climate change. Generally, refugee settlements are found to be especially vulnerable to climate change impacts through systematic geographic isolation in environmentally vulnerable locations. Furthermore,  restrictive local asylum policies are found to reinforce disenfranchisement of refugees and often limit their abilities to cope with changing environmental conditions \cite{VanDenHoek2021,Blair2020}. This vulnerability to climate change makes the need to understand refugee-environment relationships even more pressing. In this context, understanding the dynamics between refugee settlement establishments and land management practices such as the use of fire can add further insights into the environmental impacts of refugee settlements. While these insights are required for better planning and management as well as ensuring the sustainability of host and refugee relationships, they could provide an interesting perspective on potential reinforcing cycles of specific climate change vulnerability of refugee settlements and further environmental degradation of their hosting regions. 

BA maps represent the spatial extent of an area that experienced fire between two distinct points in time \cite{Giglio2018}. While fires have a historic, often positive, role to play in the long-term health of ecosystems \cite{FAO2006}, they can significantly impact the local environment and negatively influence local climate resilience through post-fire effects such as soil erosion, interruptions of hydrological regimes, changing vegetation compositions and in extreme circumstances loss of life and destruction of property \cite{Szpakowski2019}. 

\section{Related Work}
The advancements of remote sensing data and analysis methods for BA mapping have transformed global research and policy-making communities' capabilities to assess global and local fire dynamics, trends and outbreaks in a near-real-time manner \cite{Szpakowski2019}. For example, the MODIS-based MCD64A1 collection 6 model is providing global and daily BA assessments for researchers, policy-makers, and fire managers \cite{Giglio2018}. However, the 500m spatial resolution of this model leads to a high detection threshold for BA and thus, small BAs are often omitted in the product\cite{Roteta2019}. Since smallholder agricultural practices (e.g., slash and burn) are the predominant driver of fires and BA in Uganda and much of Sub-Saharan Africa, the high detection threshold of MODIS-based BA models leads to the omission of up to 70\% of total BA in Sub-Saharan Africa \cite{Nakalembe2021,Zubkova2019,Roteta2019}. Possibly due to these high omission rates and the complexity of fire practices, a recent study of refugee-environmental relationships found no statistical evidence for an increase in the MODIS-based BA Index MCD43A4 in the surroundings of refugee settlements in 49 African countries \cite{Maystadt2020}.

With limited abilities to capture fire dynamics in the surroundings of refugee settlements in Northern Uganda with current state-of-the-art BA models, there is a need to develop automated and scalable BA models with higher spatial resolution and lower BA detection thresholds for these areas \cite{Roteta2019}. While recent advances in the use of supervised deep learning (DL) models have brought significant improvements to BA detection thresholds and provide a growing knowledge base for building efficient and scalable BA models with higher spatial resolution data (e.g. Sentinel-2 remote sensing data) \cite{Pinto2020,Knoop2020}, the lack of BA ground-truth data in Sub-Saharan Africa remains a limiting factor for using those supervised DL methods in these critical regions.

\section{Proposed Methodology and Preliminary Results}
To overcome the discussed challenges with existing BA models, we propose to train a DL BA model on a region with ample ground-truth data and subsequently apply that model to a region with little to no ground-truth data through transfer learning. With the high availability and quality of BA ground-truth data in Portugal and a local presence of semi-arid climatic conditions and vegetation cover, we propose to train a DL framework on Portuguese BA data and 20m spatial resolution Sentinel-2 remote sensing data from 2016 (fire-record year in Portugal) and then apply this model to a range of refugee-hosting districts in the West Nile region in Northern Uganda from 2015-2020. See Table \ref{tab:districts} for a proposed list of refugee settlements, districts, refugee populations and settlement establishment dates.
By obtaining high-resolution BA mappings in those districts we aim to analyze how the establishment of refugee settlements possibly influenced the local fire dynamics.

 \begin{table}
  \begin{tabular}{cccc}
    \toprule
    Settlement&District&Total refugees &Established\\
    \midrule
    Rhino Camp &Arua &116,374 &1980\\
    Imvempi &Arua &64,486 &02/2017\\
    Bidi Bidi &Yumbe &231,395 &08/2016\\
    Lobule & Koboke &5,393 &09/2013\\
    Palorinya &Moyo &122,238 &12/2016\\
    Adjumani  &Adjumani &212,710 &2014-2018\\
  \bottomrule
\end{tabular}
\caption{Selected refugee settlements in West Nile, districts and establishment dates. The Adjumani settlement represents a total of 18 smaller settlements in Adjumani district \cite{UNHCR_FS2020}}
\label{tab:districts}
\end{table}

\begin{figure}
    \centering
    \includegraphics[width=0.4\textwidth]{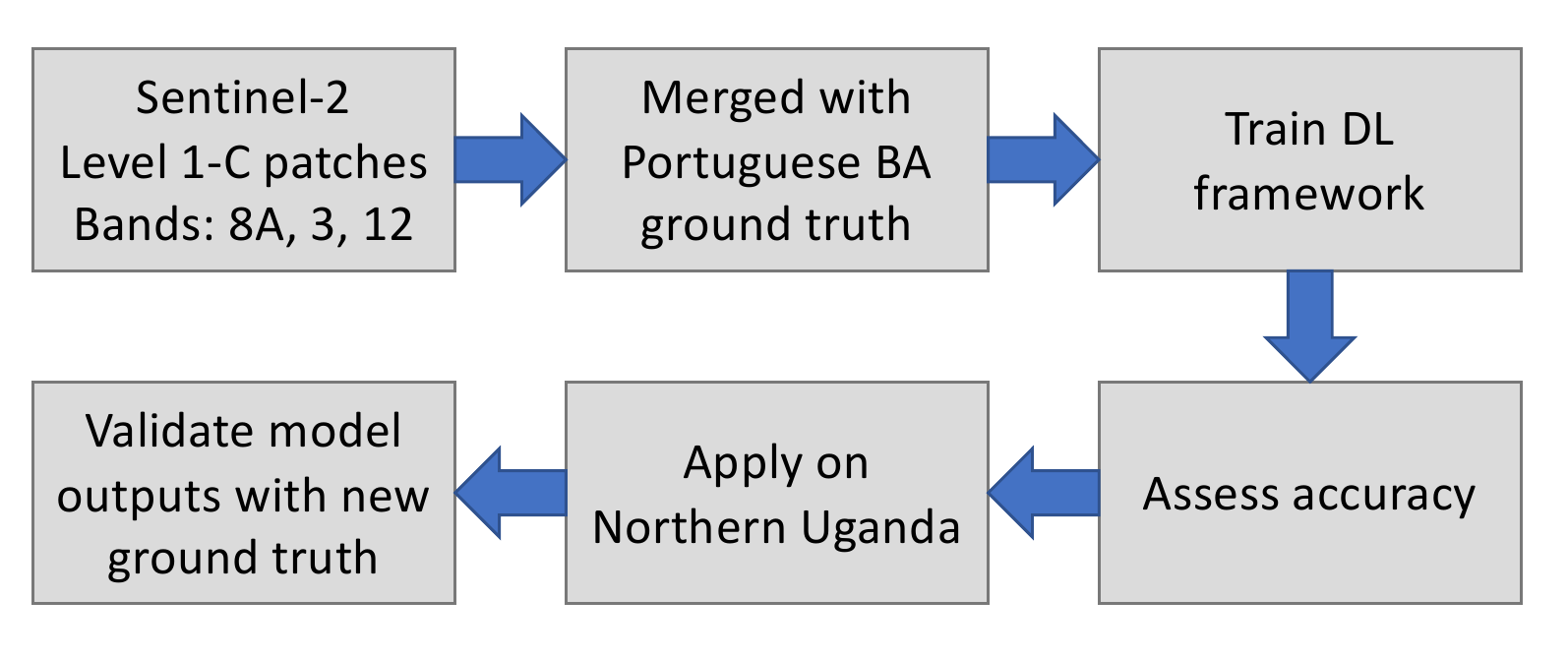}
    \caption{Overview of the key steps of the proposed methodology}
    \label{fig:overview_method}
\end{figure}

Our approach consists of six key steps, also outlined in Figure \ref{fig:overview_method}:
\begin{enumerate}
    \item Merge the Near-Infrared (8A), Green (3) and Shortwave-Infrared (12) bands, often used to highlight BA \cite{Riebeek2014}, from the Sentinel-2 Level-1C data product for Portugal (2016) and the selected West Nile districts (2015-2020).
    \item Extract $128\times128$-pixel patches from the merged false-color images and stack the patches with the Portuguese BA ground-truth labels (available at: \cite{ICNF2017}).
    \item Train the DL model on the Portuguese ground-truth data.
    \item Assess the accuracy of the model outputs with Intersection-over-Union (IoU) and Dice scores.
    \item Apply the model for inference on the selected West Nile districts and the West Nile region as a whole (as control).
    \item Evaluate the model outputs with a hand-labeled validation dataset from Northern Uganda using IoU and Dice metrics.
\end{enumerate}

While the proposed steps are work in progress, steps 1 to 4 have been already conducted and step 5 is in experiment. After merging the bands to raster images, we extracted patches that overlapped with BA ground-truth within a 90-day timeframe and thus collected a total of 2,704 128x128 false-color images that were split into 70\% training/validation and 30\% test images. We used a U-Net \cite{Ron15} architecture that includes convolutional layers in an encoder and decoder network. We used ResNet-18, an 18-layer residual neural network for the encoder and a decoding PSPNet \cite{Zhao2017} for the decoder. We used a batch size of 16 for training. The model output is a pixel-level semantic segmentation map with two classes: burned and not burned.

\begin{figure}
    \centering
    \includegraphics[width=0.4\textwidth]{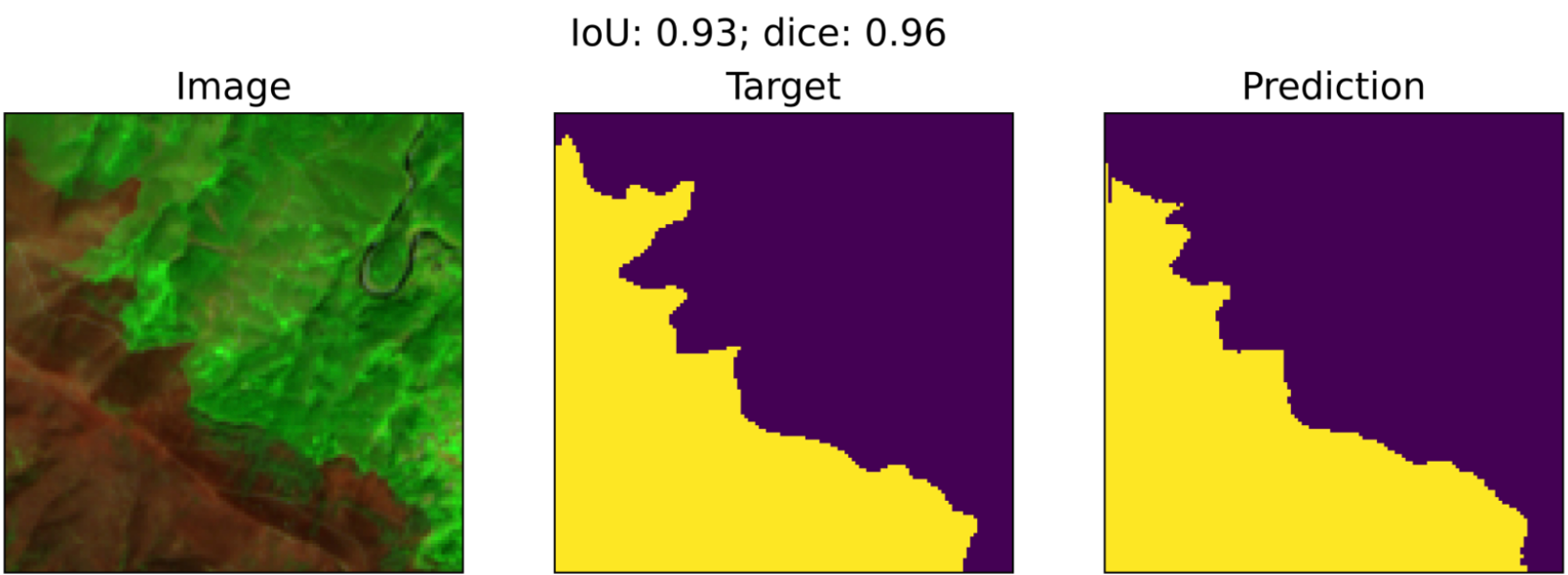}
    \caption{Testing example: Clearly visible BA as dark red in the NIR, Green, SWIR false-color image (left), matching ground-truth map (middle) and model output (right). IoU and Dice scores signal the high similarity between ground-truth and model output.}
    \label{fig:val_results}
\end{figure}

The accuracy of the predicted segmentation maps was assessed with IoU and Dice scores. We obtained averaged IoU of $0.559 \pm 0.3$ and Dice score of $0.661 \pm 0.3$ on the test dataset, as presented in Figure \ref{fig:val_results}.

\begin{figure}
    \centering
    \includegraphics[width=0.4\textwidth]{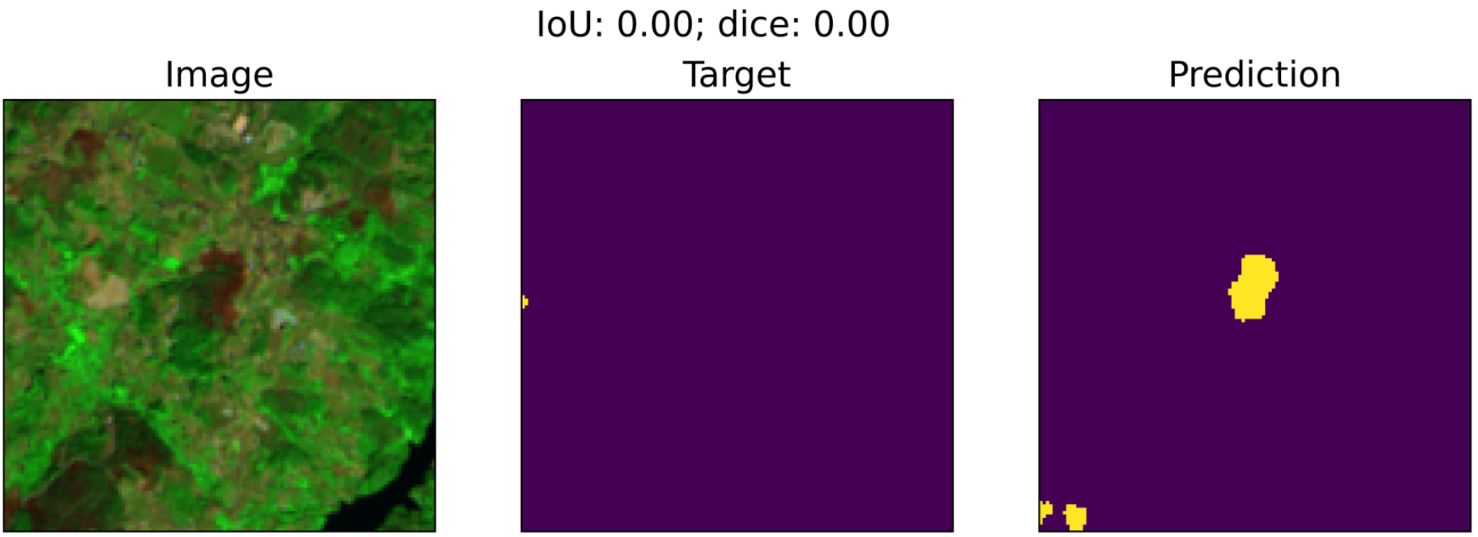}
    \caption{Example of an error in the ground-truth data (middle) that omits, along others, a clearly visible BA in the center of the patch (left), leading to IoU and Dice scores of 0.}
    \label{fig:val_errors}
\end{figure}

While the BA model showed good performance for most of the Portuguese scenes, we observed several inaccuracies in the ground-truth data that resulted in lower IoU and Dice scores for some samples. Figure \ref{fig:val_errors} shows an example input for which the model correctly predicted the presence of burned area that was not captured in the ground-truth label. This suggests that there might be other label errors in the training set that the model was able to overcome during training. To more accurately assess test performance in future work, we will visually validate test samples to remove samples with potential label errors. 

\begin{figure}
    \centering
    \includegraphics[width=0.4\textwidth]{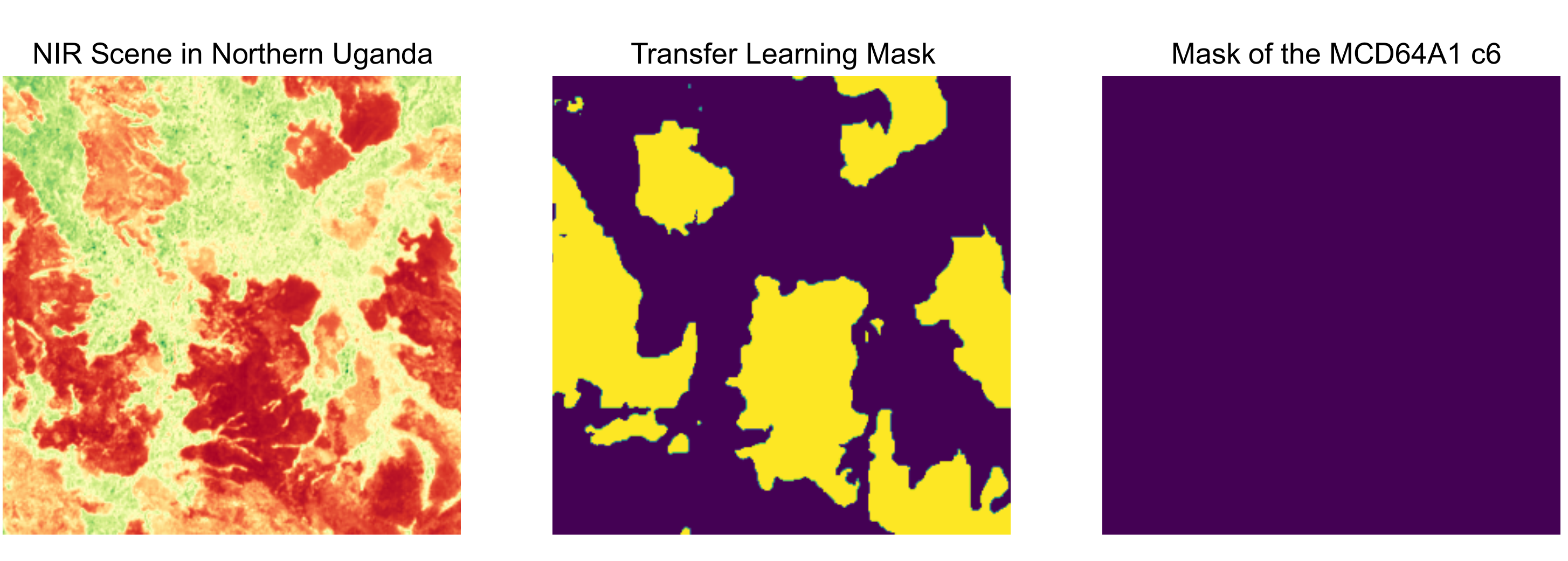}
    \caption{First set of applications of the transfer learning model to the West Nile region. Visual comparison of the NIR-only false-color image (left), the corresponding transfer learning mask (middle), and the MCD64A1 mask (right) demonstrate the significant difference between the two models.}
    \label{fig:Experiment_Uganda}
\end{figure}

First experimental transfer applications of the trained BA model on Northern Uganda showed a large amount of additionally detected BAs when compared with the MCD64A1 output for the same scene (see figure \ref{fig:Experiment_Uganda}. While a validation with actual ground-truth will be crucial to determine whether detected BA artifacts in the Uganda scene correspond to actual BA, we can already detect a significant difference between our model and the MCD64A1 BA output (which essentially did not detect any BA in the present scene). With a large-scale application of the transfer learning approach to the proposed districts and regions, we will further be able to investigate the challenges that might arise from the significant domain shift between Portugal and Northern Ugandan vegetation. Further biases and accuracy limitations might also be induced by using the atmospherically uncorrected Sentinel-2 Level 1C data product that was used due to the lack of an atmospherically corrected Sentinel-2 product (Level 2) for the year 2016 on the Sentinel HUB.

To ensure reproducibility of the conducted steps and promote future research on BA mapping, we also include our GitHub repository (\url{https://github.com/prhuppertz/Burned_Area_Detection}), that includes all processing, training, testing and assessment code with specific reproduction guidance and notes on parameter choices, such as hyperparameters of the DL model.

\section{Overall Vision}
The goal of this work is to help better estimate the impact of refugee settlement establishments on local fire dynamics and thus deliver new insights on the complex refugee-environment relationships. While the focus of our study is the West Nile region in Northern Uganda, we hope this approach can be applied and further developed by other researchers to study the refugee-BA relationships in other regions, as well. Furthermore, this work will assess the potential to create globally scalable BA models using a DL model trained with data from a region with ample ground-truth data to directly infer BA in an area with little to no ground-truth data.

\begin{acks}
Authors thank Jevgenij Gamper and the Cervest team for the significant help in building and documenting the data processing and DL training pipeline used and published in this paper.
\end{acks}
\bibliographystyle{ACM-Reference-Format}
\bibliography{refugees_BA}
\end{document}